# Multi-output LSTM soft sensors for industrial anaerobic co-digestion using LSTM-beta-VAE generative data augmentation

Rohit Murali[a,b], Benaissa Dekhici[a,d], Tao Chen[a,b], Dongda Zhang[c], Michael Short[a,b,d *]

[a] *School of Chemistry and Chemical Engineering, University of Surrey, Guildford, GU2 7XH, UK*
[b] *Surrey Institute for People-Centred Artificial Intelligence, University of Surrey, Guildford, GU2 7XH, UK*
[c] *School of Chemical Engineering and Analytical Science, The University of Manchester, Manchester, M13 9PL, UK*
[d] *Supergen Bioenergy Impact Hub, Energy and Bioproducts Research Institute, UK*

**Abstract**

Anaerobic digestion (AD) requires close monitoring of key process variables, including pH, the FOS/TAC ratio (ratio of volatile organic acids to alkaline buffer capacity) and biogas production, to maintain biological stability and prevent process failure. While mechanistic frameworks such as the Anaerobic Digestion Model No. 1 are highly parameterised, computationally expensive, and difficult to calibrate for real-time applications, data-driven Long Short-Term Memory (LSTM) networks offer a potentially more adaptable alternative. However, current data-driven approaches in AD are predominantly limited to single-output architectures that focus solely on production yields, as opposed to simultaneous prediction of vital stability indicators. Furthermore, industrial AD datasets are typically sparse and skewed toward steady-state operations. Baseline evaluations demonstrated this scarcity prevents standard algorithms (Random Forest, XGBoost) and unaugmented LSTMs from recognising the transition to unstable operations, resulting in poor estimation of stability indicators ($R^2 \leq 0.26$). To overcome these challenges, this study proposes a novel generative data augmentation framework using an LSTM-beta-Variational Autoencoder coupled with a multi-output LSTM (MO-LSTM) soft sensor. By synthesising physically valid sequences of these transitions to unstable operations from one year of full-scale plant data, the method safely populates sparse regions of the feature space. Trained on this augmented dataset, the MO-LSTM achieved robust simultaneous estimation across all targets ($R^2$ = 0.87-0.89), when using 5,000 synthetic sequences. Locally adaptive conformal prediction bounded estimations within 90% coverage intervals, dynamically signalling increased uncertainty during transient operational shifts. Furthermore, a streamlined model relying exclusively on standard operational data removing offline laboratory feedstock characterisation, marginally improved predictive capabilities ($R^2$ = 0.85-0.91). This approach mitigates data sparsity, providing a basis for hybrid mechanistic-data-driven modelling and early-warning soft sensors in industrial bioprocesses.

**Keywords:** Anaerobic Digestion, Variational Autoencoder, Long Short-Term Memory, Bayesian Optimisation

## 1.0 Introduction

Anaerobic Digestion (AD) stands as a cornerstone technology in the transition toward a circular bioeconomy, offering a dual solution for organic waste valorisation and renewable energy production [1]. However, the biological complexity that makes AD versatile also makes it inherently difficult to control. Industrial facilities operate in a dynamic environment characterised by fluctuating feedstock compositions, variable loading rates, and non-linear biological responses [2]. To maintain process stability and maximise biogas production, operators require predictive tools capable of anticipating system behaviour and pre-empting process inhibition [3].

Historically, the Anaerobic Digestion Model No. 1 (ADM1) has been the gold standard for simulating the AD process [4]. It uses a system of differential equations to describe physicochemical equilibria and biochemical kinetics [5]. However, the application of ADM1 at full-scale industrial sites faces significant hurdles. The model typically involves over 26 dynamic state variables and eight algebraic variables, requiring the definition of more than 100 stoichiometric and kinetic parameters [5]. Determining these parameters, such as half-saturation constants, maximum growth rates and biomass decay rates, requires real-time measurement of all model states, which is rarely feasible in practice. Consequently, model calibration typically relies on a limited set of readily available measurements, meaning the resulting parameters often act as empirical fitting factors rather than retaining their strict physical meaning [5]. Furthermore, ADM1 requires inputs to be fractionated into carbohydrates, proteins, and lipids where in real-world plants processing heterogeneous feedstock composition fluctuates daily, making continuous, accurate input characterisation not possible with low uncertainty [6]. Consequently, mechanistic models such as the ADM1 are often too computationally expensive and stiff to adapt to the high-frequency noise and unmeasured disturbances characteristic of industrial sites [7].

This rigidity and calibration burden have driven a shift toward data-driven modelling, to learn system behaviour directly from historical plant data without requiring explicit knowledge of underlying biological constants [8]. Machine Learning (ML) algorithms have gained significant attention in recent studies focusing on AD [3]**.** Compared to physics-based models, ML-based approaches offer several advantages, including easy implementation through open-source tools, high generalisability, uncertainty quantification, and fast execution times, making them particularly suitable for practical applications [9]. Research has recognised the benefits of ML methods and has applied them to various aspects of AD, including prediction, optimisation, and the identification of key parameters [10]. ML models can map inputs to outputs without requiring an understanding of the underlying physical processes [11].

Data-driven soft sensing is attractive because it learns directly from historical and real-time plant data, capturing nonlinearities that are difficult to represent mechanistically. Algorithms such as support vector machine (SVM), artificial neural networks (ANN) and random forest (RF) have been applied in AD applications for predictions showing promising results [3,9]. While these models perform well on steady-state data, they struggle to represent dynamic transients because they treat each observation as independent [12,13]. Multi-task Gaussian processes have been applied to AD [14] but[14] performed well only when the training and testing conditions were similar. Among ANN models, Long short-term memory (LSTM) is a deep learning algorithm that can retain valuable information from the past in its memory [11]. An LSTM network is a form of recurrent neural network (RNN), which was developed to solve the vanishing gradient problem of RNNs by retaining or discarding the data over long sequences. LSTMs are well suited to retaining long-term dependencies in sequential data and have been shown to improve the accuracy of methane production predictions [13,15,16].

Despite the established success of LSTMs in capturing temporal dependencies, a fundamental limitation persists within the current literature. Existing research is heavily dominated by Single-Output architectures, focusing almost exclusively on predicting methane or biogas production [11,13,17]. While maximising energy recovery is an economic imperative, this single-dimensional focus neglects the multivariate nature of process stability. A robust system requires the simultaneous prediction of production outputs alongside critical stability indicators, which offer early warnings of instability, to fully capture the coupled dynamics between loading rates and the risk of inhibition [18]. To the best of the authors' knowledge, however, the joint LSTM estimation of production and stability indicators has not previously been reported for full-scale AD.

An important barrier to developing these robust Multi-Output models is the small data problem in AD. Industrial AD datasets are typically long but narrow and contain years of high-frequency Supervisory Control and Data Acquisition (SCADA) data for basic parameters (temperature, pH and biogas flow) along with feedstock quantities but have low sampling frequencies for the critical stability indicators which are obtained via offline laboratory analysis due to the high capital cost of online sensors [19]. This mismatch poses a severe challenge for standard ML approaches [20]. Furthermore, historical data tend to consist almost entirely of steady-state operations as operators actively avoid process failures. ML algorithms, however, require large and diverse datasets to learn the complex boundaries between stable and unstable operating regimes. In the absence of sufficient examples of process deviations, models are prone to overfitting and fail to generalise beyond the limited training data, instead capturing site-specific noise rather than the underlying biological dynamics [21]. To address these limitations, there has been significant interest in soft sensors as a cost-effective alternative, where ML models representing the system can help infer difficult-to-measure variables from easy-to-measure online data [22].

## 1.1 Challenges in Synthetic Data Generation

Initial efforts addressed the challenge of small datasets by generating simulated data using established mechanistic models, such as the ADM1 [8]. However, these models require extensive calibration kinetic parameters and often fail to capture the high-frequency noise and unmeasured disturbances characteristic of real industrial sensors. To overcome this, statistical augmentation methods have been successfully adopted in the AD literature. Notably, [23] and [24] used Random Standard Deviation Sampling (RSDS) to expand limited datasets by injecting Gaussian noise based on statistical variance, while others have employed the Synthetic Minority Over-sampling Technique (SMOTE) [13] to address class imbalance. While these approaches have demonstrated success in improving the prediction of specific steady-state parameters, building dynamic time-series soft sensors introduces structural challenges. Additionally, given the nonlinear and dynamic nature of biogas production, the use of methods such as SMOTE is generally not recommended, since SMOTE operates by linearly interpolating between existing samples in a static feature space, ignoring temporal sequences [25]. Consequently, interpolating between a steady state from one event and a process failure state from another does not necessarily yield a chemically valid transition state, as the augmentation mechanism itself ignores the fundamental reaction kinetics. Therefore, to safely expand training data for real-time process monitoring, the chosen augmentation method must explicitly understand and preserve the correct chronological sequence of events.

Alternatively, Generative Adversarial Networks (GANs) have also been explored for data augmentation in AD, helping to expand limited datasets and identify optimal operating conditions [26]. However, GANs are notoriously difficult to train and often suffer from mode collapse, where the generator produces only a limited variety of samples, failing to capture the full distribution needed to represent rare process deviations or instability events [27]. The use of Variational Autoencoders (VAEs) in AD has been proven effective for soft sensing validation [28]. Shaw et al. [29] advanced the field by applying beta-Variational Autoencoders (β-VAEs) for accurate downstream predictive modelling for complex substrates such as Palm Oil Mill Effluent, demonstrating the immense value of VAE-based augmentation. Inspired by their effective framework, the present study seeks to extend this generative approach. While the β-formulation improves latent space disentanglement, these prior architectures typically rely on static feed-forward layers, treating data points as independent vectors [29]. Consequently, they often fail to capture the long-term temporal dependencies and dynamic evolution of biological states essential for process monitoring. This limitation can be addressed by LSTM-VAEs, which have been applied to anomaly detection outside AD [30]. However, their potential for generative time-series augmentation in AD remains unexplored.

### 1.2 Research Objectives and Novelty

This paper addresses critical gaps in the existing literature by proposing a novel, integrated deep learning framework. Rather than solely applying existing algorithms, this study converges the challenges of data scarcity and stability monitoring to deliver site-ready predictions. Specifically, the core novel contributions of this work are:

1. *Temporal Augmentation via LSTM-β-VAE*: We introduce a novel sequence-based generative pipeline designed specifically for AD data. Existing studies combining data augmentation with LSTM prediction often rely on tabular generation methods that treat data points as independent vectors. These static approaches generate isolated synthetic samples that may not represent chemically valid transition states. Our framework addresses this limitation by integrating an LSTM encoder-decoder directly into the β-VAE architecture. This allows the model to synthesise intact, high-fidelity temporal windows that preserve both the long-term dependencies and the complex physicochemical correlation structures of real site data, as evidenced by the rigorous statistical fidelity analysis in Section 3.2.

2. *Joint estimation of Production and Stability Indicators*: While existing literature predominantly focuses on single-output predictions (e.g., isolating biogas or methane yield), AD is governed by deeply coupled biochemical interactions. To address this, we propose a Multi-Output LSTM architecture to simultaneously predict biogas production alongside critical biological stability indicators (pH and FOS/TAC) with and without the data augmentation. By leveraging multi-task learning, this approach captures the coupled dynamics of the process and improves predictive robustness over standard single-target models, providing a holistic early-warning framework.

3. *Dispensability of Complex Feedstock Characterisation*: We demonstrate whether the predictive framework can operate effectively without relying on time-consuming, offline laboratory analyses of feedstock characteristics (Section 3.3). We establish that a streamlined model relying exclusively on standard operational parameters maintains, and marginally improves, predictive capabilities, significantly reducing the data-acquisition burden for industrial deployment.

The remainder of this paper is structured as follows: Section 2 outlines the methodology for synthetic data generation and multi-output LSTM model development. Section 3 covers the experimental results, consisting of baseline model predictions using purely site data, the validation of synthetic data fidelity, and the effectiveness of the augmented data for training the multi-output LSTM. Finally, Section 4 summarises the key findings, acknowledges study limitations, and outlines future directions.

## 2.0 Methodology

This section details the development of the proposed framework, including data acquisition, the architecture of the generative model (LSTM-β-VAE), and the predictive model (Multi-Output LSTM).

### 2.1 Data Acquisition and Preprocessing

This study uses historical operational data from a full-scale UK AD site operating continuous anaerobic co-digestion (ACoD). The system comprises a primary digester, followed by a post-digester in series, with total volumes of 3,328 $m^3$ and 3,533 $m^3$, respectively. The plant processes a heterogeneous mixture of agricultural residues, primarily consisting of baseload Maize Silage and Rye Silage, supplemented by three seasonal variants: Grass Silage, External Maize Silage, and External Grain Maize.

The dataset consists of two full years of operational data. Data acquisition frequencies varied significantly by variable type. Standard operational parameters including feedstock quantities, process water addition, hydraulic retention time (HRT), temperature, organic loading rate (OLR), mixer power and current, and flaring hours were recorded at high frequencies via the plant's online Supervisory Control and Data Acquisition (SCADA) system. Conversely, critical stability indicators (pH and FOS/TAC) and specific feedstock characteristics (Total Solids, Volatile Solids, $NH_4$-N, and Total Kjeldahl Nitrogen) were obtained via offline laboratory analysis, leading to lower sampling frequencies and inherent data gaps. A complete statistical summary of all input features, operational limits, and missing data proportions, is presented in Table 1, with detailed feedstock characteristics provided in Supplementary Materials Appendix A, Table A1.

**Table 1:** Statistical summary of historical plant data and input features.

| Category | Feature/ Variable | Mean | Standard Deviation (SD) | Min | Max | Missing Data (%) |
|---|---|---|---|---|---|---|
| **Operational Inputs** | OLR in Fresh Matter basis (kg FM/$m^3$/d) | 36.44 | 5.95 | 3.74 | 40.96 | 0 |
| | HRT (d) | 55.36 | 12.29 | 34.95 | 139.82 | 0 |
| | Process Water ($m^3$) | 17.74 | 10.14 | 0 | 24.03 | 0 |
| | Temperature (°C) | 48.12 | 0.83 | 46.2 | 50.3 | 0 |
| | Flaring (hours) | 0.59 | 3.25 | 0 | 24.03 | 0 |
| | Gas roof (%) | 67.8 | 12.9 | 44 | 100 | 0.3 |
| | Mixer Power (kW) | 5.02 | 0.5 | 3.5 | 6.13 | 0 |
| | Mixer Current (A) | 11.15 | 0.52 | 9.92 | 12.3 | 0 |
| **Feedstock Inputs** | Total Feed (tonnes) | 110 | 17.96 | 11.3 | 123.7 | 0 |
| | Maize Silage (tonnes) | 80.43 | 31.62 | 8.2 | 123.7 | 0 |
| | Rye Silage (tonnes) | 14.04 | 11.12 | 0 | 37.54 | 0 |
| | Grass Silage (tonnes) | 2.58 | 3.29 | 0 | 10.56 | 0 |
| | External Maize Silage (tonnes) | 11.53 | 18.39 | 0 | 74.68 | 0 |
| | External Grain Maize (tonnes) | 1.46 | 2.25 | 0 | 6.97 | 0 |
| **Target Outputs** | Total Biogas ($m^3$) | 27400 | 4645 | 4772 | 31930 | 0 |
| | pH | 7.62 | 0.13 | 6.90 | 7.99 | 58.7 |
| | FOS/TAC | 0.31 | 0.08 | 0.11 | 0.63 | 58.2 |

## 2.2 Generative Model: LSTM-β-VAE

The proposed generative model employs an LSTM-β-VAE architecture to learn the underlying probability distribution of the multivariate AD process, enabling the generation of high-fidelity synthetic temporal sequences. The architecture of the generative model is shown in Figure 1.

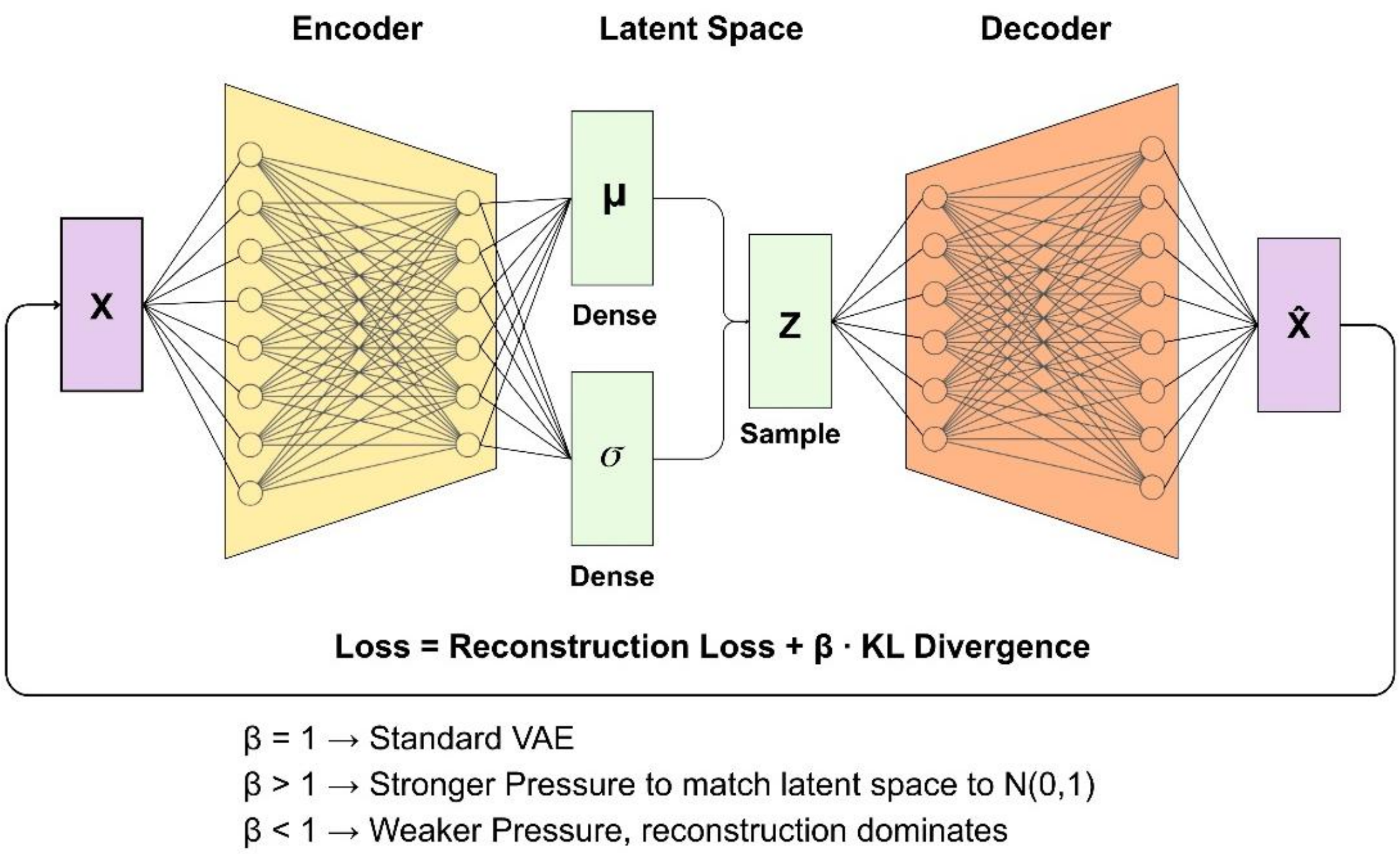

Figure 1 - Schematic of LSTM-β-VAE architecture including encoder, latent space and decoder.

To capture complex temporal dependencies, the standard VAE dense layers are replaced with a stacked LSTM encoder. Based on preliminary architectural screening, the encoder depth was fixed at two layers (256 and 128 units), and the decoder mirrored this capacity (128 and 256 units). Both used a Tanh activation function by batch normalisation and dropout (rate 0.3) to prevent overfitting. This encoder maps input sequences $X \in \mathbb{R}^{T \cdot F}$, with $T = 30$ time steps and $F = [n]$ variables, to a compressed latent distribution [30]. The final hidden states are passed through two parallel fully connected layers to predict the mean ($\mu$) and log-variance ($\log \sigma^2$) of a diagonal Gaussian posterior $q_\phi(z|X)$. To generate new sequences, the model samples from this space. A reparametrised Gaussian latent space allows for continuous sampling and backpropagation via $z = \mu + \sigma \odot \epsilon$, where σ = exp($\frac{1}{2}$ log $\sigma^2$) and $\epsilon \sim \mathcal{N}(0,1)$ and $\odot$ denotes element-wise multiplication [31]. This isolates the stochasticity so that gradients can propagate through the sampling step. The decoder mirrors the encoder's structural capacity, and the sampled latent vector $z$ is repeated $T$ times, concatenated with the input at each step, and processed by a stacked LSTM (128 and 256 units) to reconstruct the sequence $\hat{X}$.

To bias reconstruction towards the variables of greatest operational interest while maintaining temporal fidelity, the network optimises a composite $\beta$-VAE objective [29]. This objective balances the overall fidelity of the generated sequences against the regularisation of the latent space. The overarching total loss ($\mathcal{L}_{\text{Total}}$) is defined in Eq.1 as:

$$\mathcal{L}_{Total} = \mathcal{L}_{Recon} + \beta \cdot D_{KL}\left(q_\phi(z|X)||p(z)\right) \quad \text{(Eq. 1)}$$

where $p(z) = \mathcal{N}(0, \mathrm{I})$ and the Kullback-Leibler divergence term $D_{KL} = -\frac{1}{2}\sum_k(1 + \log \sigma_k{}^2 - \mu_k{}^2 - \sigma_k{}^2)$, summed over the latent dimensions and averaged over the batch. The coefficient $\beta$ controls the trade-off between reconstruction fidelity and the strength of latent regularisation. To counteract the temporal over-smoothing typical of VAE reconstructions and to prioritise the stability indicators, the reconstruction term ($\mathcal{L}_{\text{Recon}}$) is defined as:

$$\mathcal{L}_{Recon} = MSE(X, \hat{X}) + \omega \cdot MSE(X_C, \hat{X}_C) + \gamma[MSE(\Delta X, \Delta\hat{X}) + \omega \cdot MSE(\Delta X_C, \Delta\hat{X}_C)] \quad \text{(Eq. 2)}$$

where $C$ is the set of three critical outputs (Total Biogas, pH and FOS/TAC), $X_i$ is the $i$-$th$ variable of the sequence, and $\Delta X_t = X_t - X_{t-1}$ for $t = 2, \ldots, T$ is the first-order temporal difference. Each MSE is averaged over the batch and time steps, and the full-sequence terms are also averaged over all F variables. Since each critical variable contributes to both the full and the variable-specific terms, its effective weight relative to a non-critical variable is $1 + \omega F$. With $\omega = 3$ (fixed) and $F = [n]$, the critical outputs therefore receive approximately $[1 + 3n]$ times more emphasis.. The coefficient $\gamma$ controls the contribution of the temporal-difference terms, which penalise mismatched step-to-step dynamics.

The architecture was fixed, while the latent dimensionality and loss coefficients were tuned by Bayesian optimisation using Optuna with a Tree-structured Parzen Estimator [32] over 50 trials. To find the optimal trade-off between sequence reconstruction accuracy and latent space flexibility, the hyperparameter search space was explicitly bounded. The latent dimensionality ($d_z$) was evaluated within a discrete range of [20, 60] in steps of 5 to efficiently capture meaningful architectural differences. Simultaneously, the disentanglement penalty ($\beta$) and the volatility weight ($\gamma$) were tuned on continuous log scales within the bounds of [0.05, 0.5] and [0.5, 5.0], respectively. Since the purpose of the VAE is generative rather than predictive, each trial was scored on the fidelity of its output rather than on reconstruction loss alone. After training, each trial generated 1,000 sequences from the prior, which were inverse-scaled and used solely to compute the trial score. For the critical variables, this score averaged the agreement between the synthetic and real standard deviations, the agreement between their ranges, and the agreement between their step-to-step volatility, with the volatility term weighted twice. A penalty of 0.1 times the lowest training loss reached was then subtracted. Each trial was trained with the Adam optimiser (learning rate $5 \times 10^{-4}$, batch size 32) for up to 200 epochs, with early stopping on the training loss (patience 20) and median pruning. The final model was retrained with the selected hyperparameters for up to 500 epochs (patience 50), retaining the weights with the lowest training loss, and was then used to generate the final synthetic dataset of 5,000 sequences (Section 2.3). No separate validation partition was used for the VAE, instead its generalisation was assessed through the statistical and temporal fidelity analysis in Section 3.2.

Prior to training, the raw sensor data was pre-processed to address temporal inconsistencies. As established in **Table 1**, measurements for critical stability indicators (pH and FOS/TAC) were recorded roughly three times per week, representing a true measurement frequency of approximately 41%. To enable continuous sequence-based training for the LSTM encoder, linear interpolation was applied to bridge these intermittent gaps and preserve general trend continuity [20]. However, to ensure rigorous evaluation and prevent the artificial inflation of performance metrics by trivially predictable smooth interpolated lines, all downstream test metrics were computed exclusively against true laboratory-measured data points only.

To facilitate stable network convergence, the dataset was scaled. To strictly prevent data leakage, all scaling parameters were fitted exclusively on the training data partition prior to any model exposure. Standard operational variables were transformed using a standard scaler. Conversely, the three critical variables (Total Biogas, pH, and FOS/TAC) were transformed using a robust scaler on the median and interquartile range (IQR) to minimise the influence of extreme transient outliers inherent to volatile AD systems. To further enrich the training set with rare transitional behaviour, each 30-day window was scored on two complementary measures computed over the critical variables. The first was a transience score, the largest single-step change within the window, which captures sudden process upsets. The second was an extremity score, the largest deviation from the window's own median, which captures sustained drifts into unusual operating regions. Each score was normalised by its maximum value and the two were summed, and windows at or above the 75th percentile of this combined score were added two further times to the training set.

## 2.3 Synthetic Data Generation Strategy

Once the generative model was fully trained on the limited site data, it was used to generate a large synthetic dataset comprising 5,000 independent temporal sequences. This framework produces complete temporal windows simultaneously by sampling directly from the learned latent distribution.

The generation process followed a three-step protocol: (1) Latent Space Sampling: A set of 5,000 random latent vectors was drawn from a standard normal prior distribution, $\mathcal{N}(0,1)$. The dimensionality of these vectors corresponded to the optimal latent size ($d_z$) established during the hyperparameter tuning phase. Since the VAE was trained on a dynamically balanced dataset where sequences exhibiting high step-to-step volatility were explicitly oversampled the learned latent space was inherently structured to dedicate a larger proportional volume to these transient states. Consequently, sampling directly from the standard normal prior naturally yielded a synthetic dataset enriched with the sparse, unstable operational conditions required to robustly train the downstream predictive model. (2) Sequence Decoding: The sampled latent vectors were passed through the trained decoder network. The decoder projected each compressed vector back into the temporal domain, simultaneously outputting 5,000 synthetic sequences. Each generated sequence spanned a full temporal window of $T = 30$ time steps across all sensor features, capturing the underlying multivariate system dynamics. (3) Robust Inverse Transformation: To ensure the synthetic sequences accurately reflected real-world physical magnitudes, an inverse scaling procedure was applied to the decoder's raw outputs. To maintain absolute fidelity to the preprocessing stage, variables were inverted using the exact transforms fitted to them during training. Standard sensor features were reverted to their original scales using the inverse functions of the standard scalers. Conversely, the critical operational outputs (Total Biogas, pH, and FOS/TAC) were reverted using a robust inverse scaling method based on their original medians and interquartile ranges.

2.4 Predictive Model: Multi-Output LSTM

The predictive model architecture, shown in Figure 2, is formulated as a dynamic soft sensor. Rather than relying on autoregressive past measurements of the targets, the model maps purely operational inputs to biological system states. It processes a rolling multivariate historical window of 30 time steps at daily intervals ($X_{t-30:t}$), consisting of the feed characteristics and operating conditions. The past values of the three target variables (Total Biogas, pH, and FOS/TAC) are explicitly excluded from the input feature matrix. By forcing the network to predict the plant's stability purely from precursor operational data, the architecture structurally prevents persistence forecasting. Instead, the model learns the complex, non-linear interactions between feedstock variations and impending biological responses.

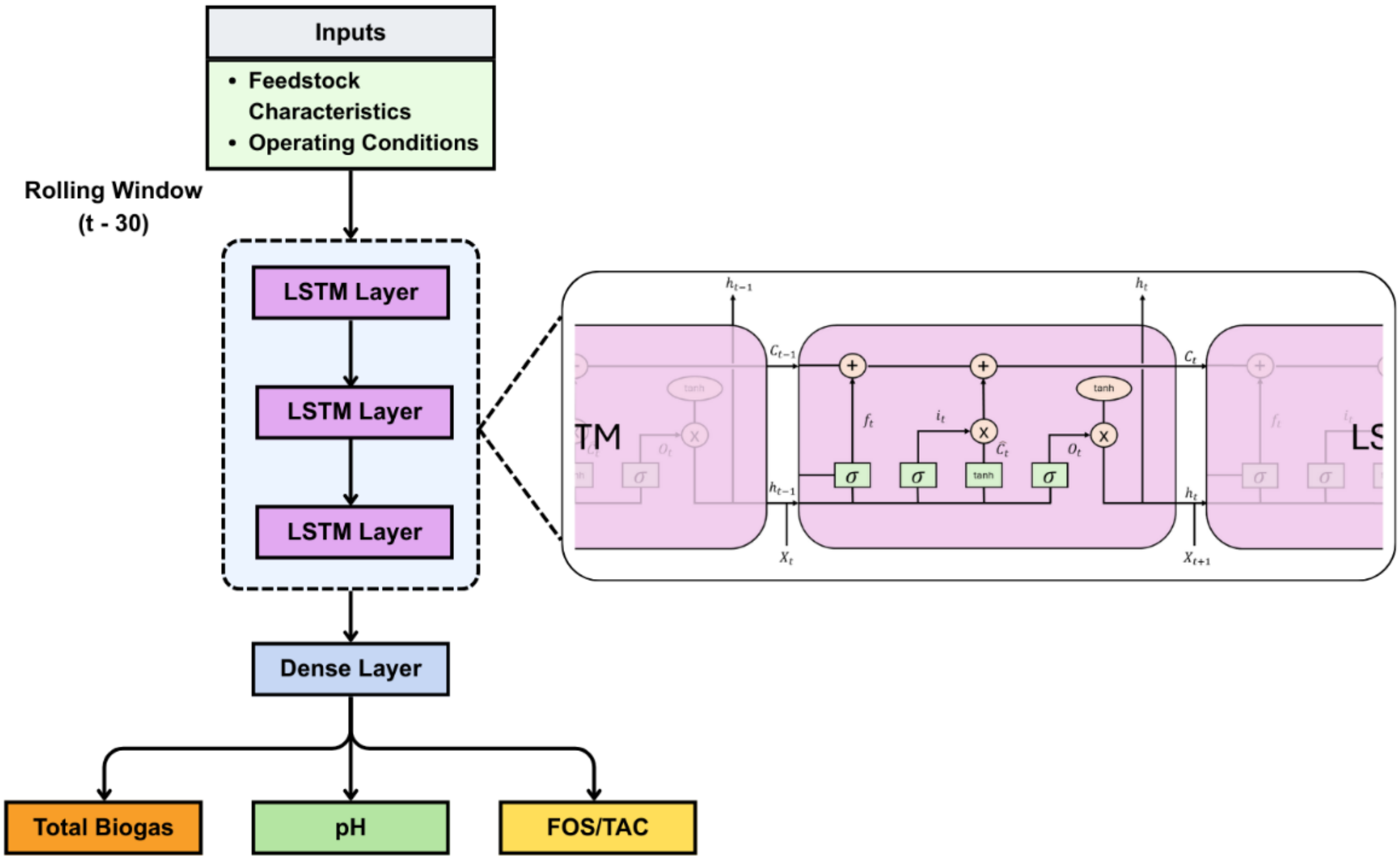

Figure 2 - Multi-Output LSTM overall model architecture including inputs of feedstock characteristics and operating conditions and predicting outputs including total biogas, pH and FOS/TAC.

The network uses a stacked LSTM framework, dynamically optimised via Optuna to contain between one and three hidden layers (exploring capacities up to 256, 128, and 64 units respectively), alongside dropout layers (0.1 to 0.5) to prevent overfitting. The final layer is a fully connected dense layer with three neurons and a linear activation function, mapping directly to the concurrent target variables at the end of the temporal window. A shared Mean Squared Error (MSE) loss is optimised across all three targets simultaneously. This multi-target objective ensures that the network extracts generalised features relevant to the stability of the entire AD system, rather than overfitting to the variance of a single isolated output.

Finally, to assess sensitivity to random initialisation, the final model with the Optuna-selected hyperparameters was trained five times with random seeds, keeping the data split, preprocessing and hyperparameters fixed. Each run used early stopping on validation loss (a patience of 30 epochs) with learning-rate reduction on plateau. We report the mean and standard deviation of each metric across the five runs, as well as the metrics of the ensemble obtained by averaging the five models' predictions. Since pH and FOS/TAC were not measured daily, all metrics were computed on measured days only.

### 2.5 Baseline Comparative Framework

Before evaluating the proposed framework using synthetic data, a comprehensive baseline was established to demonstrate the inherent limitations of standard algorithms when applied to sparse historical AD datasets. These baselines serve to isolate the specific benefits of the LSTM-β-VAE's generative augmentation. The comparative framework consisted of four distinct approaches:

1. *Naive and Persistence Forecasts:* To establish absolute predictive floors, both a naive forecast (predicting the static mean of the training targets) and a Last-Observation-Carried-Forward (LOCF) persistence model was implemented. These confirm whether the AD process is too dynamically non-stationary to be predicted by simple statistical averages or by simply carrying the last offline laboratory measurement forward.
2. *Linear Time-Series:* A multi-output Ridge Regression was applied to the flattened historical rolling window to evaluate whether standard linear combinations of the inputs could adequately capture stability indicators.
3. *Tabular Machine Learning:* A multi-output Random Forest (RF) and XGBoost regressors were used to establish a strong, non-linear, decision-tree-based benchmark.
4. *Fixed-Architecture LSTM:* To verify that deep learning failures on the raw dataset are fundamentally due to data sparsity rather than poor model conditioning during the hyperparameter search, a standard multi-output LSTM using a sensible, fixed architecture (128 and 64 units) was evaluated.

To ensure a strictly equitable performance comparison, all baseline models were subjected to the exact same chronological data segregation and evaluation metrics as the final proposed framework, detailed in Section 2.6.

### 2.6 Experimental Design and Data Segregation

To rigorously evaluate the hypothesis that synthetic data augmentation improves predictive robustness, a strict data segregation approach was enforced to completely prevent data leakage and look-ahead bias. The baseline historical dataset was partitioned chronologically into three distinct sequential periods: a training set, a validation set, and test set. The LSTM-β-VAE was fitted exclusively to the chronological training partition with zero exposure to the subsequent validation or test periods. This ensured that the 5,000 generated synthetic sequences encoded only past information without leaking future temporal dynamics.

Unless otherwise stated, the multi-output LSTM was subsequently trained exclusively on the 5,000 synthetic sequences generated by the β-VAE using a year's long of operational data. This isolated the parameter-update phase from the specific noise and operational anomalies present in the limited historical data. The non-training real data was partitioned into validation and testing sets using a strictly chronological 50/50 split. The real validation set (the period immediately following the training data) was used to monitor performance and trigger early stopping, ensuring that only network weights that generalised to real-world plant operation were rewarded. The remaining real data formed a strictly chronological, held-out test set at the very end of the time series, which was used only once for final evaluation.

Model optimisation was again achieved by Bayesian optimisation via the Optuna framework. To ensure a strictly equitable performance comparison, both the historical baseline model and the synthetic-augmented model were subjected to the exact same tuning approach, consisting of 50 Optuna trials exploring identical hyperparameter search spaces. For both models, the Bayesian tuning strategy systematically explored the architectural depth (1 to 3 hidden layers), LSTM unit capacity (32 to 256 units), dropout regularisation (0.1 to 0.5), and learning rates ($10^{-5}$ to $10^{-2}$ on a log scale). The objective function for both models was explicitly designed to minimise the MSE on the exact same real validation and test data. Early stopping was applied with a patience of 30 epochs, strictly monitored against the real validation loss. Simultaneously, automated model checkpointing was used to capture and save the exact network weights that maximised real-world generalisation.

To determine how much synthetic data is needed, the MO-LSTM was also trained on subsets of 366, 500, 1,000, 2,500 and 5,000 synthetic sequences, all drawn from the same trained LSTM-β-VAE. The smallest subset (366 sequences) matches the number of daily windows in one year of real training data, First, the hyperparameters selected for the full model were held fixed, which isolates the effect of data volume. Then Optuna was re-run for each subset size (30 trials), so that each size received its own best architecture. Each configuration was trained with five random seeds, using the same real validation and test partitions as Section 2.6.

## 2.7 Model Validation and Performance

Model predictions were evaluated against the measured site data using the following metrics.

a) Root Mean Squared Error (RMSE):

The Root Mean Squared Error (RMSE) measures the average magnitude of the prediction errors, providing a direct indication of absolute model accuracy in the exact units of the target variable. It is heavily weighted toward penalising larger estimation variances. As presented in Eq.1, this is calculated using the total number of observations in the dataset ($n$), the observed site value at time step $i$ ($y_{sim,i}$), and the predicted value at time step $i$ ($y_{obs,i}$):

$$RMSE = \sqrt{\frac{1}{n}\sum_{i=1}^{n}(y_{sim,i} - y_{obs,i})^2} \quad \text{(Eq.1)}$$

b) Coefficient of Determination ($R^2$):

The Coefficient of Determination ($R^2$) was calculated to evaluate the proportion of variance in the observed data that is successfully captured by the model as presented in Eq.2. This metric indicates the goodness-of-fit relative to the mean of the observed values ($\bar{y}_{obs,i}$):

$$R^2 = 1 - \frac{\sum_{i=1}^{n}\left( y_{obs,i} - y_{sim,i}\right)^2}{\sum_{i=1}^{n}\left(y_{obs,i} - \bar{y}_{obs,i}\right)^2} \quad \text{(Eq.2)}$$

c) Mean Absolute Percentage Error (MAPE):

The MAPE provides an intuitive measure of the prediction accuracy, expressed in Eq. 3 as a percentage deviation of the predictions from the observed values:

$$MAPE = \frac{100\%}{N}\sum_{i=1}^{N}\left|\frac{y_{sim,i} - y_{obs,i}}{y_{obs,i}}\right| \quad \text{(Eq.3)}$$

While MAPE is highly effective for quantifying the relative magnitude of the error for variables like biogas flow and FOS/TAC, it is less meaningful for interval-like, logarithmic quantities such as pH. Consequently, for pH evaluation, absolute RMSE (expressed in pH units) is prioritised to allow for direct comparison against standard industrial sensor measurement uncertainty.

All simulations were performed on a workstation equipped with an AMD Ryzen 7 3700X processor (8 cores, 16 threads, 3.6 GHz) and 32 GB RAM. The computational environment was built on Python 3.11, with deep learning architectures implemented via TensorFlow 2.20.0 and data preprocessing executed using scikit-learn 1.7.2. Hyperparameter optimisation was conducted using Optuna 4.4.0. Comprehensive details regarding global random seed initialisation, total Optuna trials, epoch configurations, batch sizes, and model training times are documented in the Supplementary Materials Appendix B.

## 3.0 Results and Discussion

This section presents and discusses the experimental results of the proposed framework. First, a comprehensive suite of baseline models consisting of naive forecasts, linear regression, tabular machine learning is evaluated to highlight the inherent limitations of learning from sparse historical datasets. Subsequently, the generative capabilities of the LSTM-β-VAE are analysed, followed by an evaluation of the augmented Multi-Output LSTM to demonstrate the impact of synthetic sequence generation on soft sensor accuracy.

### 3.1 Baseline Model Performance

To establish a rigorous comparative baseline and demonstrate the inherent limitations of training predictive models on sparse historical datasets, a set of baseline algorithms was evaluated exclusively on the real data. Full details of the model structures along with the MO-LSTM training can be found in the Supplementary Material Appendix B, Table B1 and Figure B1. First, a naive forecast using the training-set mean (Train Mean) and a persistence forecast, which carries the last observed value forward, were evaluated. The persistence forecast completely failed to capture the biological stability indicators on the held-out test set, yielding highly negative coefficients of determination ($R^2$ = -0.99 for pH, $R^2$ = -0.81 for FOS/TAC). This confirms that the AD process is highly dynamic, and stability cannot be predicted by simply carrying offline laboratory measurements forward. Next, linear (Ridge Regression) and non-linear tree-based machine learning regressors (Random Forest and XGBoost) were applied to the flattened rolling windows. While XGBoost emerged as the strongest conventional baseline for mapping broad volumetric trends, achieving an $R^2$ of 0.51 for Total Biogas, its performance severely degraded on the critical stability indicators ($R^2 = 0.22$ for pH, $R^2 = -0.08$ for FOS/TAC). Similarly, Random Forest achieved an $R^2$ of 0.46 for Total Biogas but failed to capture the complex biological dynamics of FOS/TAC ($R^2 = 0.07$). This demonstrates that powerful standard tabular machine learning models, despite their capacity to map general gas production, overfit the steady-state historical data and struggle to generalise to the non-linear boundaries of process failure.

To verify that these limitations extend to deep learning architectures, the Multi-Output LSTM was evaluated on the raw historical data under two configurations: a sensible fixed-architecture baseline (2 layers, 128 and 64 units) and an Optuna-optimised configuration. This dual approach was necessary to confirm that the failure of the deep learning model was driven by inherent data sparsity rather than poor hyperparameter tuning. The fixed-architecture LSTM achieved marginal predictive capacity for Total Biogas ($R^2 = 0.23$) but failed entirely on the critical stability indicators, with $R^2$ of 0.04 for pH and $R^2$ of -0.28 for FOS/TAC. When the network was subsequently subjected to Bayesian hyperparameter optimisation via Optuna, the search algorithm applied heavy regularisation (dropout rate of 0.397) to counter the sparsity. Despite this rigorous tuning, the Optuna MO-LSTM still showed severe structural overfitting.

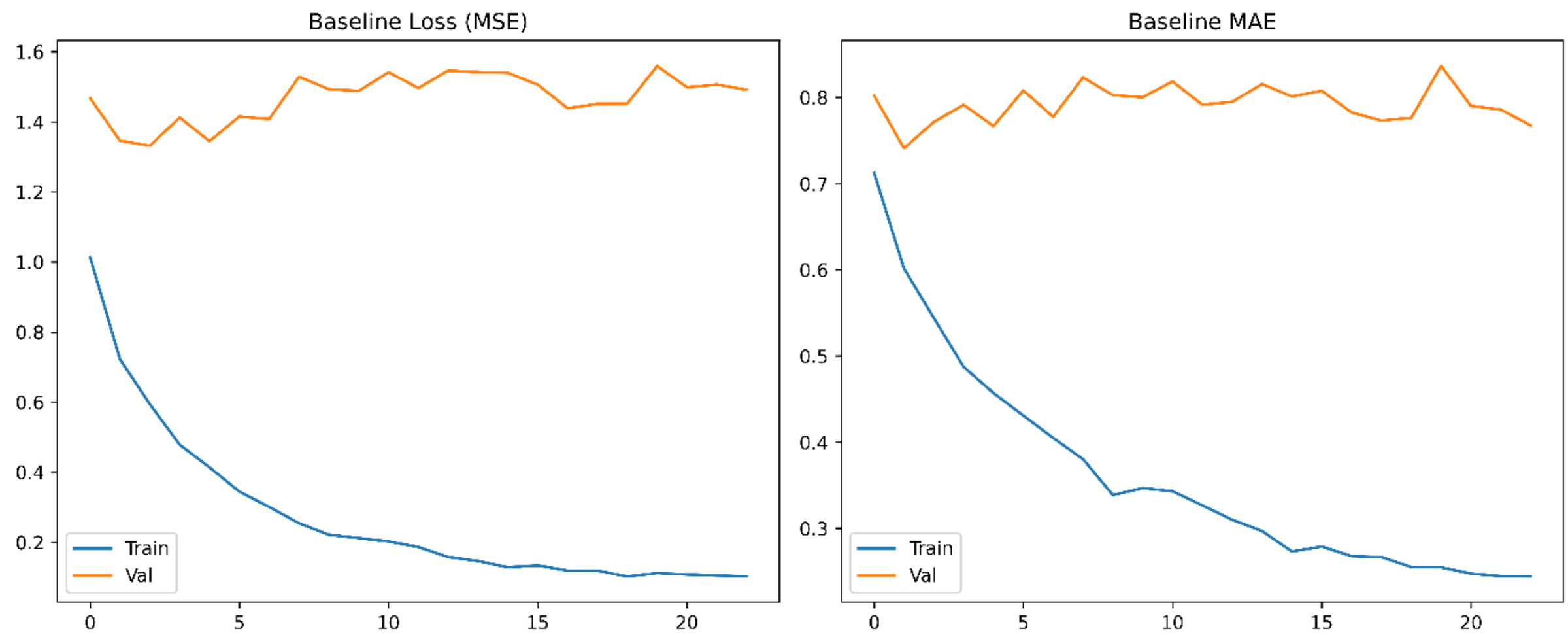

Figure 3 - Baseline model training dynamics. The divergence between the training and validation curves illustrates severe overfitting, as the network fails to generalise beyond the limited historical training data.

As illustrated in Figure 3, the training dynamics demonstrate a clear divergence: the training loss (MSE) and mean absolute error (MAE) decreased steadily, whereas the validation loss remained entirely stagnant and disconnected from the training progress. This confirms that standard LSTMs regardless of how conservative they are architected or rigorously they are tuned, they cannot learn generalisable physicochemical dynamics from this sparse historical data. While the Optuna based MO-LSTM managed a marginal predictive capacity for Total Biogas ($R^2 = 0.52$), it yielded near-zero correlation for pH ($R^2 = 0.13$) and a negative coefficient for FOS/TAC ($R^2 = -0.1$).

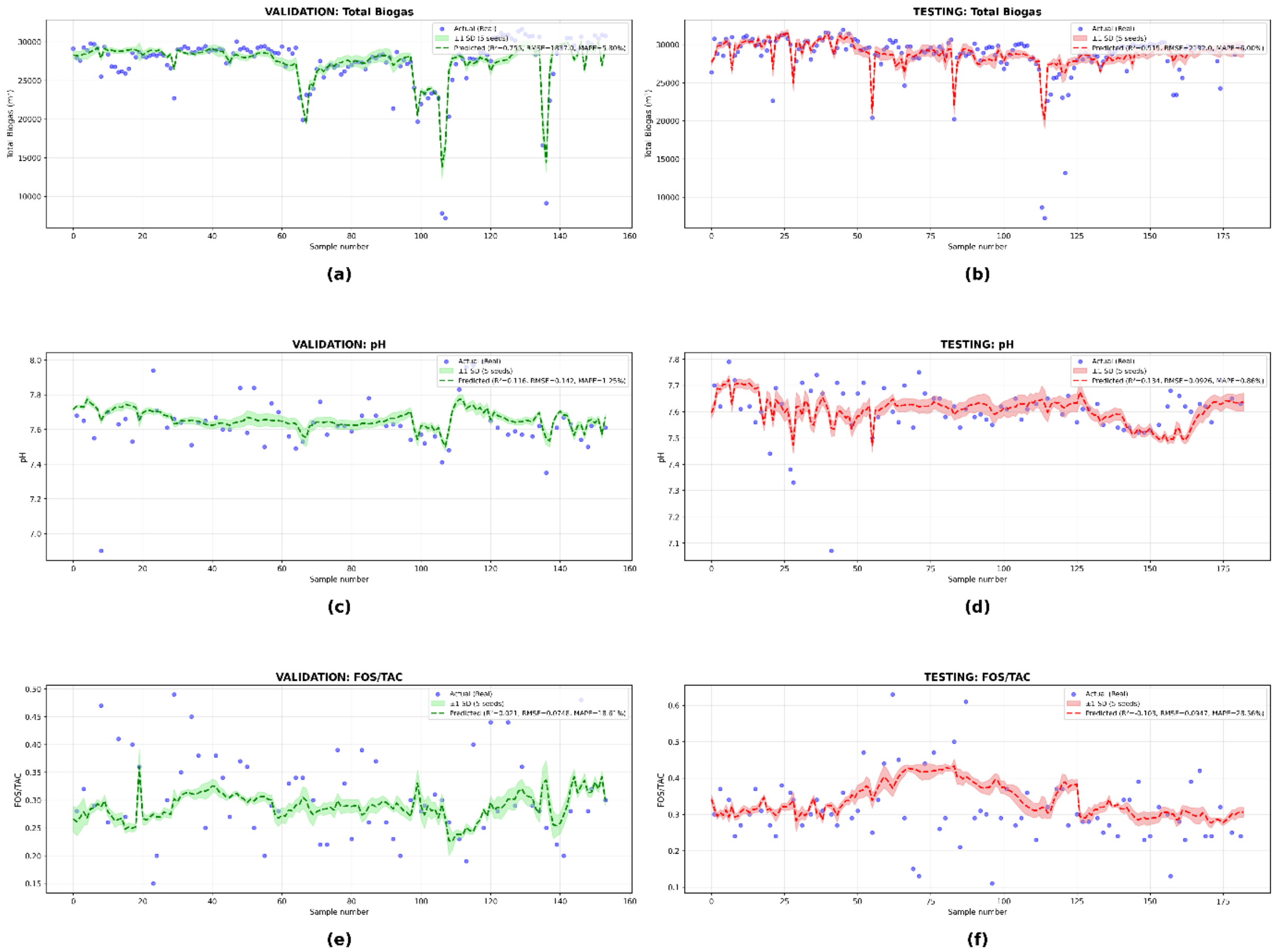

Figure 4 - Baseline Multi-Output LSTM predictions for (a, b) total biogas production, (c, d) pH and (e, f) FOS/TAC on the validation (left, green) and test (right, red) periods following Optuna hyperparameter optimisation on unseen test data. Trained exclusively on limited historical data, the model struggles to capture transient deviations. The dashed line is the ensemble mean of five models trained with five different random seeds, and the shaded band shows ±1 SD across the five seeds. Markers show measured values where pH and FOS/TAC were measured on only a subset of days, so predictions are shown daily while markers appear only on sampling days.

A summary of all baseline performances on the unseen test set is presented in **Table 2**. These results empirically confirm that standard algorithms struggle to learn the boundaries between stable and unstable operation from sparse AD datasets.

**Table 2:** Comparative performance of baseline predictive models on the unseen test set.

| **Baseline Model** | **Total Biogas** | | | **pH** | | | **FOS/TAC** | | |
|---|---|---|---|---|---|---|---|---|---|
| | **RMSE** | **$R^2$** | **MAPE (%)** | **RMSE** | **$R^2$** | **MAPE (%)** | **RMSE** | **$R^2$** | **MAPE (%)** |
| Naïve Forecast | 3181 | -0.02 | 9.48 | 0.102 | -0.06 | 0.88 | 0.090 | 0 | 23 |
| Persistence Forecast | 2951 | 0.12 | 6.68 | 0.140 | -0.99 | 1.12 | 0.1204 | -0.81 | 30.4 |
| Ridge Regression | 2750 | 0.24 | 8.42 | 0.098 | 0.02 | 0.86 | 0.094 | -0.09 | 27.7 |
| Random Forest | 2315 | 0.46 | 6.63 | 0.086 | 0.26 | 0.76 | 0.086 | 0.07 | 23.6 |
| XGBoost | 2102 | 0.51 | 6.21 | 0.088 | 0.22 | 0.81 | 0.093 | -0.08 | 25.3 |
| MO-LSTM (Fixed-Architecture) | 2752 | 0.23 | 7.96 | 0.097 | 0.04 | 0.93 | 0.101 | -0.28 | 27.4 |
| MO-LSTM (Optuna based Architecture) | 2192 | 0.52 | 6.0 | 0.093 | 0.13 | 0.86 | 0.095 | -0.10 | 28.6 |

Collectively, these baseline experiments demonstrate that regardless of model complexity from naive persistence through to a rigorously tuned deep learning architecture that the limited quantity of real historical data prevents standard models from learning the boundary between stable and unstable digester operation. While the tree-based tabular regressors (Random Forest and XGBoost) marginally outperformed the unaugmented MO-LSTM on the historical data, all baseline models ultimately failed to achieve acceptable predictive accuracy for the critical stability indicators.

The LSTM architecture was explicitly selected as the predictive foundation for this study because it natively processes three-dimensional temporal sequences, preserving the chronological reaction kinetics of the digester. Conversely, tabular models require rolling windows to be flattened into independent variables, destroying sequential autocorrelation. This structural advantage of LSTMs explains why previous studies have consistently reported them outperforming Random Forest models in AD applications [33,34]. The critical transition states leading to process instability are simply too sparse in the historical data to be captured by standard linear regression, non-linear tabular models, or non-data augmented models. To overcome this data scarcity without violating underlying biochemical kinetics, a sequence-based generative may be helpful. The subsequent sections detail how the proposed LSTM-β-VAE framework successfully populates these sparse operational regions to achieve robust, multi-target estimation.

### 3.2 Synthetic Data Generation via LSTM-β-VAE

The primary objective of the generative phase was to expand the limited historical dataset while strictly preserving the dynamic behaviour and physicochemical correlations of the AD process. The performance of the LSTM-β-VAE was evaluated based on its training stability, univariate statistical fidelity, and multivariate correlation retention.

### 3.2.1 Training Dynamics and Latent Space Utilisation

A common limitation of standard VAEs coupled with powerful decoders such as LSTMs is posterior collapse, where the model ignores the latent representation (z) and collapses the KL divergence to zero [35]. Figure 5 shows the training loss and confirms that the proposed β-formulation successfully prevented this issue. The KL divergence loss stabilised at a non-zero value, ensuring the latent space was actively used to encode sequence dynamics.

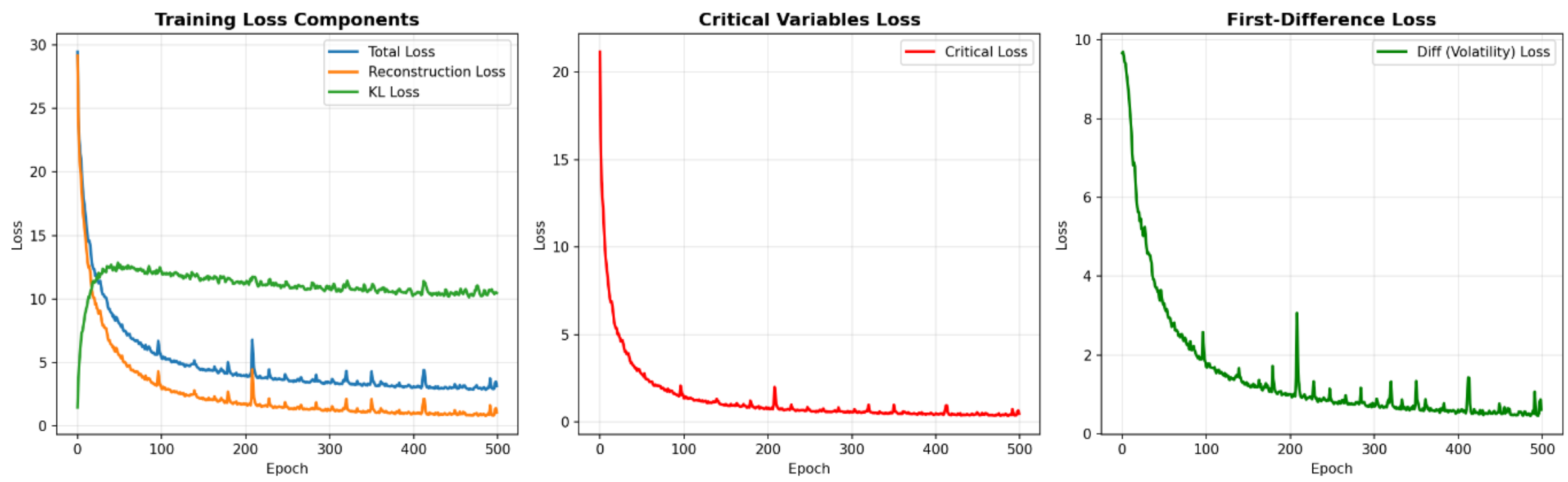

Figure 5 - Training loss components demonstrating the avoidance of posterior collapse. The non-zero KL loss confirms active use of the latent space.

Beyond preventing collapse, β also sets the trade-off between reconstruction fidelity and the strength of KL regularisation, with lower values favouring accurate reconstruction and higher values favouring a more compact, regularised posterior. The optimisation yielded a final β value of 0.20, a latent dimensionality of 25, and a volatility weight of 0.72, indicating that the search favoured reconstruction and volatility fidelity of the synthetic sequences over stronger latent regularisation.

### 3.2.2 Univariate Statistical Fidelity

To validate the generative quality at the variable level, the statistical distributions of the real and synthetic datasets were evaluated across two complementary lenses. As shown in Figure 6, Kernel Density Estimation (KDE) was first used to compare the Probability Density Functions (PDFs) and central tendencies of the critical stability indicators (Figure 6a-c). To ensure the β-VAE successfully mapped the extreme transient conditions rather than just the stable means, box plots were simultaneously employed (Figure 6d-f). This dual perspective highlights the generative model's ability to accurately capture both the core operational bulk and the volatile distribution tails and outliers inherent to the real process data.

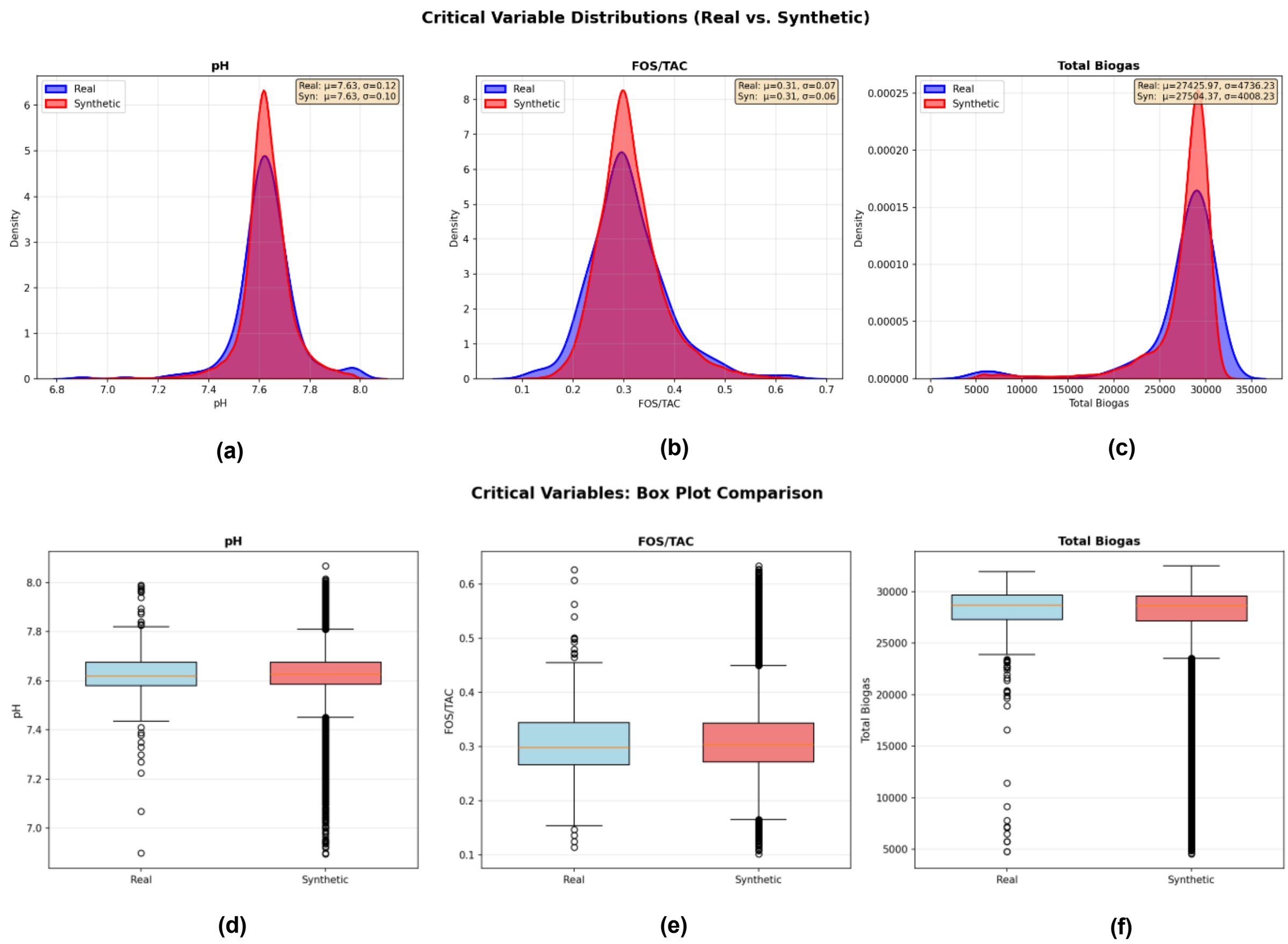

Figure 6 - Statistical comparison of real and synthetic distributions for critical process variables. The top row displays Kernel Density Estimation (KDE) plots illustrating the probability density and overlap for (a) pH, (b) FOS/TAC, and (c) Total Biogas. The bottom row provides corresponding box plots comparing the central tendency, interquartile ranges, and extreme outliers for (d) pH, (e) FOS/TAC, and (f) Total Biogas. Real historical data is denoted in blue, while VAE-generated synthetic data is denoted in red.

When dealing with raw SCADA data, extreme values inevitably comprise a mixture of genuine biological volatility and occasional sensor errors. Rather than applying arbitrary physical thresholds to artificially clean the training data which would create a distributional mismatch with the unfiltered test set and real-world deployment conditions these raw extremes were deliberately retained. By applying robust scaling (median/IQR), the network gradients were protected from being overwhelmingly skewed by absolute magnitudes, allowing the β-VAE to stably map these transient states. Furthermore, since the VAE training pipeline was intentionally designed to oversample volatile sequences, the tails of the synthetic distributions (shown in Figure 6a-c) densely represent the full spectrum of process reality. This explicitly ensures that the downstream LSTM is exposed to, and robustly regularised against, both biological shocks and inherent sensor noise.

As highlighted by both the KDE and box plots, the marginal distributions show exact or close quantitative agreement. The synthetic data means ($\mu$) deviate by less than 1% from the real data across all outputs (e.g., Real pH $\mu$ = 7.63 vs. Synthetic $\mu$ = 7.63, Real FOS/TAC $\mu$ = 0.31 vs.

Synthetic $\mu$ = 0.31; Real Total Biogas $\mu$ = 27,425 vs. Synthetic $\mu$ = 27,504). The standard deviations ($\sigma$) and absolute minimum/maximum bounds are similarly aligned, indicating that the generative model successfully reproduced both the central bulk and the realistic physical limits of the system.

To validate temporal fidelity beyond marginal distributions, the step-to-step volatility (first differences) and lagged dependencies of the sequences were evaluated.

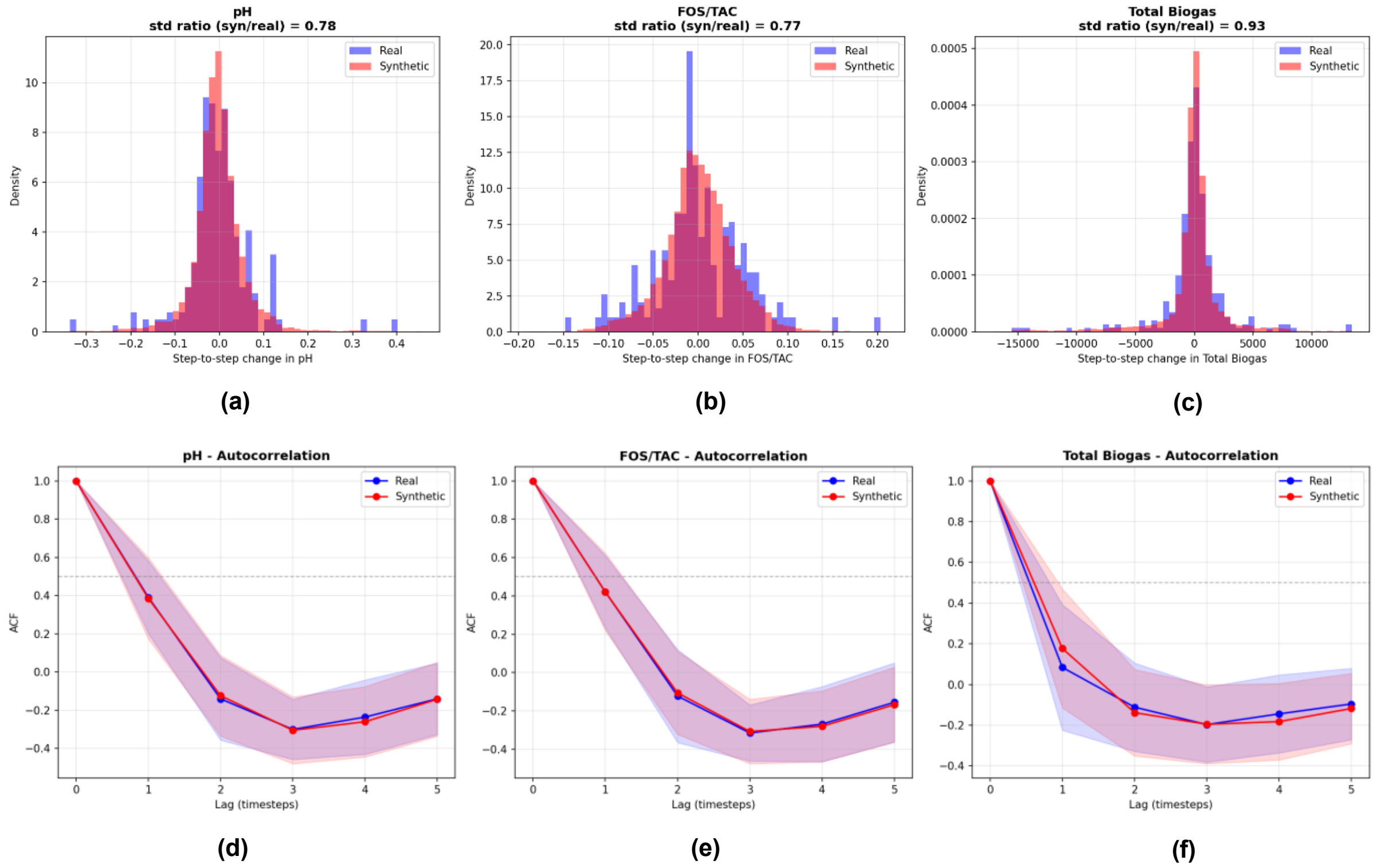

Figure 7 - Temporal fidelity and volatility validation of synthetic sequences. The top row displays the distributions of step-to-step first differences ($\Delta X_t$) for (a) pH, (b) FOS/TAC, and (c) Total Biogas, demonstrating the generative model's ability to capture realistic operational volatility. The bottom row displays the Autocorrelation Function (ACF) up to 5 timesteps for (d) pH, (e) FOS/TAC, and (f) Total Biogas, confirming the preservation of lagged temporal dependencies required for downstream LSTM training.

The standard deviations of the step-to-step changes demonstrated high parity between the real and synthetic sets, particularly for Total Biogas (Syn/Real ratio = 0.93) and well for pH (Syn/Real ratio = 0.78) and FOS/TAC (Syn/Real ratio = 0.77). Two-sample Kolmogorov-Smirnov (KS) tests on these first-difference distributions also yielded test statistics of 0.069 for pH, 0.107 for FOS/TAC, and 0.056 for Total Biogas, confirming strong structural alignment. Furthermore, autocorrelation function (ACF) analysis confirmed that the synthetic sequences successfully preserved the multi-step temporal decay and lagged dependencies inherent to the real continuous process. This rigorous temporal validation shown in Figure 7 firmly justifies the use of these synthetic sequences to train the downstream LSTM.

### 3.2.3 Preservation of Physicochemical Correlations

As established, conventional statistical augmentation methods (e.g. RSDS and SMOTE) are not suitable in biochemical applications because they treat variables independently, generating chemically invalid states. To verify that the LSTM-β-VAE overcame this limitation, the multivariate relationships of the generated data were analysed. Pearson correlation matrices computed for the real and synthetic datasets show closely similar structure (Figure 8). The strong inverse and direct dependencies between Biogas, pH and FOS/TAC are retained.

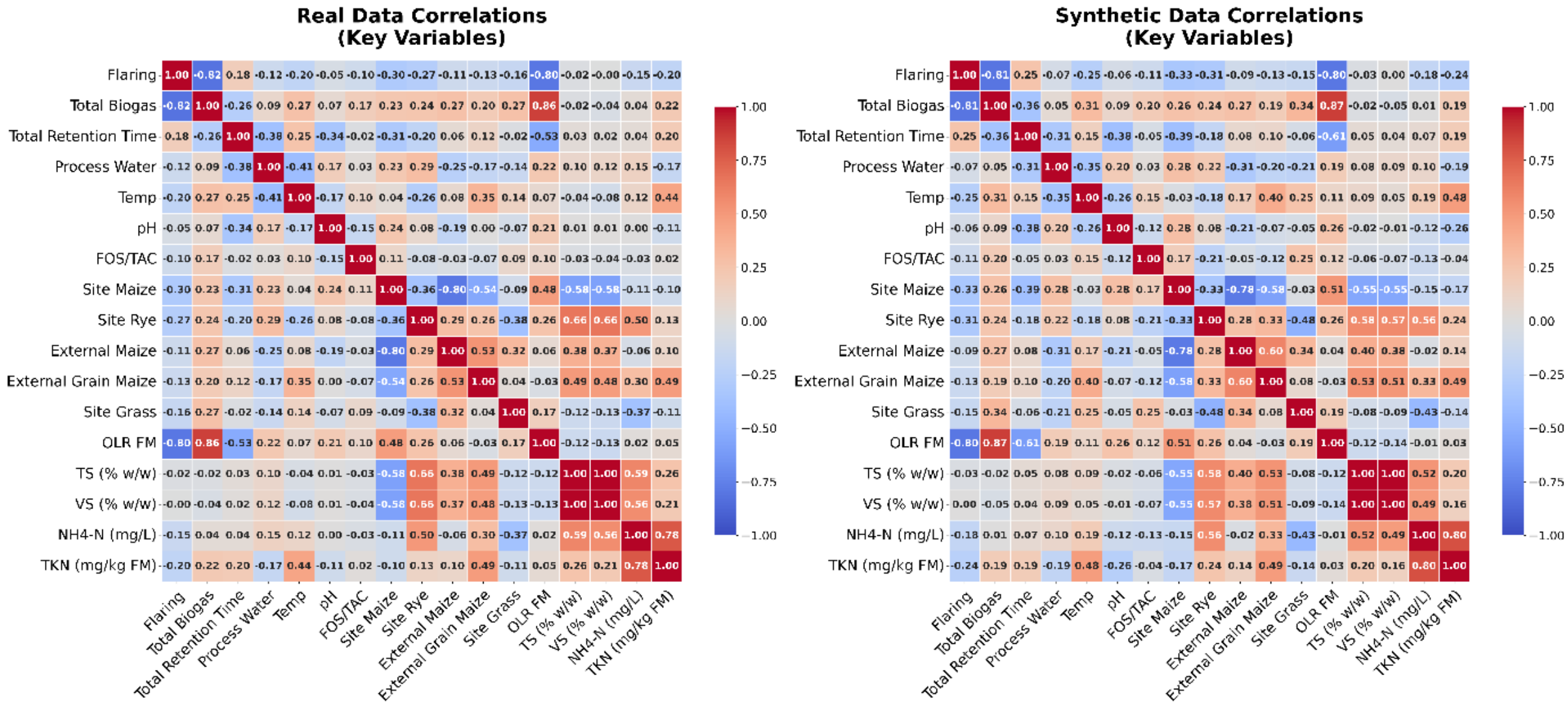

Figure 8 - Pearson correlation matrices demonstrating the preservation of multivariate dependencies in the synthetic dataset.

Rather than relying on visual inspection of the heatmaps alone, the maximum absolute element-wise difference, mean absolute difference, and Frobenius norm of the difference between the real and synthetic correlation matrices were computed. For the three output variables total biogas, pH, and FOS/TAC the correlation structure is preserved closely. The maximum absolute difference across these three variables is 0.04, the mean absolute difference is 0.019, and the Frobenius norm of the difference is 0.07 when computed over the full symmetric matrix (equivalently 0.05 over the three unique pairs, since a symmetric matrix with zero diagonal double-counts each off-diagonal pair). Individually, the total biogas-pH shifts from 0.07 (real) to 0.09 (synthetic), total biogas-FOS/TAC from 0.17 to 0.20, and pH-FOS/TAC from -0.15 to -0.12. These base correlations are themselves weak in the real data ($|r| \leq 0.17$), so while the absolute shifts are small, this reflects preservation of a weak, near-independent coupling between the three targets rather than a strong physicochemical relationship being reproduced.

Extending the comparison to the seventeen key process variables shown in Figure 8 gives a mean absolute difference of 0.041 and a Frobenius norm of 0.92, rising to 0.05 and 2.95 across the full variable set. The largest differences occur between a target variable and a minor, auxiliary variable with little influence on the process (for example FOS/TAC and Site Grass, or pH and TKN), and these were already weakly correlated in the real data. In contrast, the relationships between the three target variables themselves, and between Total Biogas and its strongest real-data drivers such as Flaring and OLR (Fresh Matter basis), are all preserved to within 0.01 to 0.04 of their real-data values.

Bivariate scatter plots (Figure 9) show that the non-linear relationships between variables, such as the inverse relationship between pH and FOS/TAC, are largely reproduced in the synthetic samples. Capturing these joint distributions is a necessary, though not sufficient, condition for the generated transition states to be physically plausible.

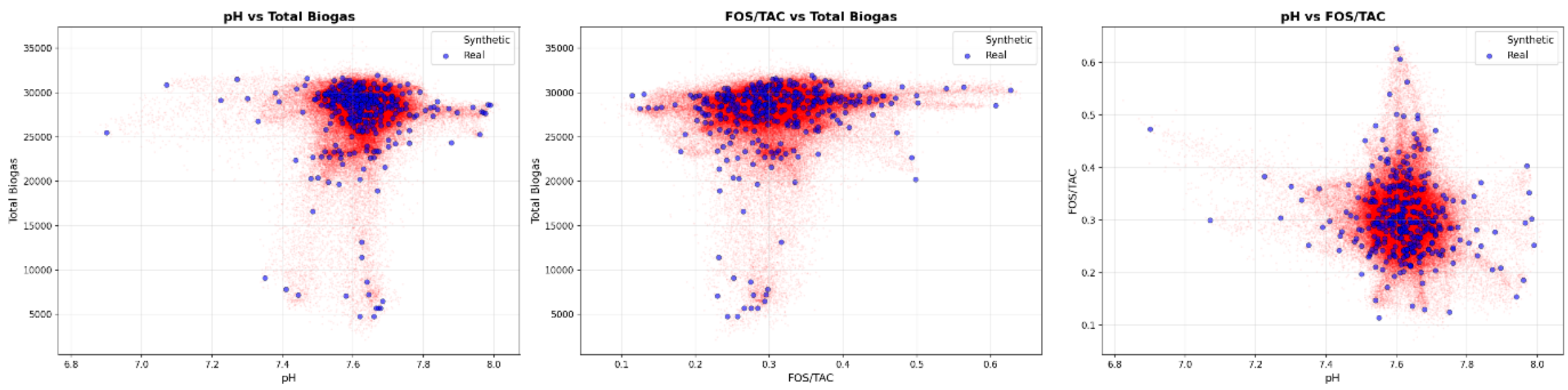

Figure 9 - Bivariate scatter plots illustrating the retention of non-linear boundaries and inverse relationships between critical variables.

Extended scatter plots analysing the relationships between additional input variables and target outputs are available in the Supplementary Material Appendix C, Figures C1-C3.

### 3.3 Multi-Output LSTM Predictive Performance

Integrating the synthetic sequences generated by the LSTM-β-VAE into the training yielded a substantial improvement across all predictive metrics. Prior to final evaluation, the augmented LSTM underwent hyperparameter tuning using the Optuna framework to ensure optimal network architecture and learning dynamics. The hyperparameter optimisation history is shown in Figure 10 with the final model training presented in the Supplementary Materials Appendix D, Figure D1-D2. The hyperparameter importances (Appendix D, Figure D3) revealed that the configuration of the deeper LSTM layers (3 layers) and the learning rate were the primary drivers of model convergence, significantly outweighing the influence of batch size or dropout rates.

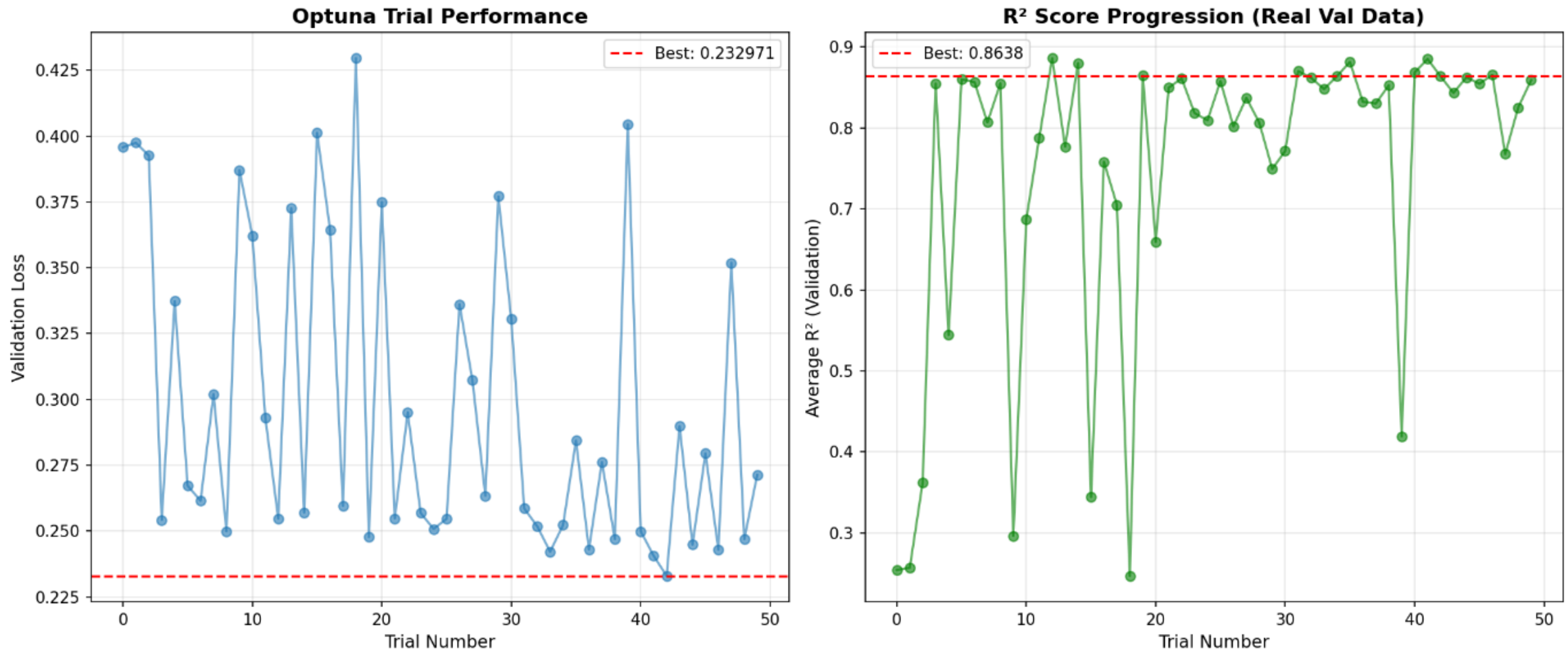

Figure 10 - Optuna hyperparameter optimisation history. The left panel displays the validation loss across successive trials, while the right panel illustrates the progressive improvement and stabilisation of the $R^2$ score, peaking at 0.86.

With the optimal configuration (presented in **Table 3**), the augmented model was evaluated across five random initialisation seeds to ensure robustness. The MO-LSTM achieved mean test $R^2$ values of 0.87 ± 0.017 for Total Biogas, 0.86 ± 0.027 for pH, and 0.89 ± 0.017 for FOS/TAC. The remarkably small standard deviations indicate that performance was highly robust to random initialisation. Averaging the five models into a final ensemble improved performance slightly across every target (test $R^2$ of 0.88, 0.87, and 0.89, respectively). By generating a large number of transitional trajectories, the LSTM-β-VAE successfully populated regions of the feature space that are sparsely represented in the historical data. This allowed the predictive LSTM to learn the associated dynamic boundary conditions rather than defaulting to the mean.

To rigorously quantify predictive uncertainty in the multi-target model, locally adaptive conformal prediction was applied [36]. Since the model's error is not constant throughout the monitoring period, prediction uncertainty was allowed to vary over time. A local difficulty measure was calculated for each day as a rolling estimate of the recent absolute prediction error (using a 14-day window). Each validation residual was normalised by this local error scale, and the 90th percentile of these normalised residuals was taken as a calibration factor. This factor was then combined with the local error scale on the test set to construct a prediction interval targeting 90% empirical coverage. To ensure the evaluation reflected only real observations, days without an actual recorded laboratory measurement for pH and FOS/TAC were excluded from the uncertainty estimation.

Figure 11 compares the resulting 90% adaptive intervals on the validation and test periods for all three targets. The ensemble achieved strong point-forecast agreement with the observed data on both periods: $R^2$ of 0.90 (validation) and 0.88 (test) for Total Biogas, 0.86 and 0.87 for pH, and 0.85 and 0.89 for FOS/TAC, with corresponding mean absolute percentage errors of 3.6%, 0.3 to 0.5%, and 7.5 to 8.9%, respectively.

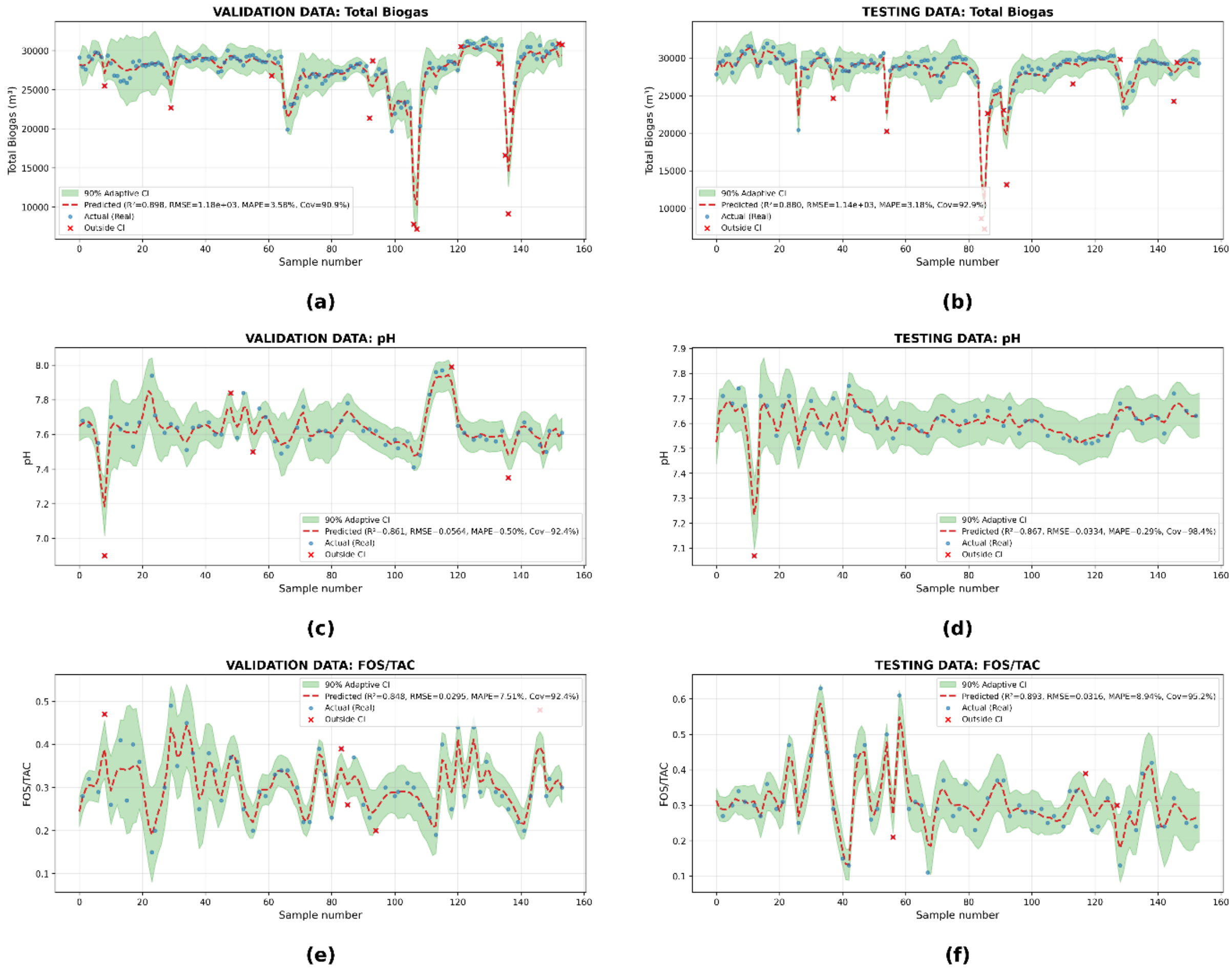

Figure 11 - Predictions of the MO-LSTM for (a, b) Total Biogas production, (c, d) pH, and (e, f) FOS/TAC on the validation (left, green) and test (right, red) periods. The dashed line is the ensemble mean of five models trained with different random seeds. The shaded band represents the 90% locally adaptive conformal prediction interval, with red crosses indicating actual observations that fell outside the predicted bounds. Since pH and FOS/TAC were subjected to intermittent offline sampling, predictions are generated daily, while markers appear only on actual sampling days.

The empirical coverage successfully met or exceeded the 90% target across all variables and both periods (90.9 to 92.9% for Total Biogas, 92.4 to 98.4% for pH, 92.4 to 95.2% for FOS/TAC), indicating that the intervals were well-calibrated rather than systematically too narrow. Only a small fraction of the true observations fell outside their predicted intervals (between 11 and 14 days out of 154 for total biogas, and between 1 and 5 days out of the 61 to 66 measured days for pH and FOS/TAC). These rare outliers (marked by red crosses in Figure 11) were strictly concentrated around the sharpest disturbances in the data, most notably in the severe dip in biogas production and the abrupt pH decline early in the test period. The prediction band width remains tight during periods of stable digester operation and expands sharply around large disturbance events. Consequently, the soft sensor successfully signals increased estimation uncertainty during transient biochemical shifts, without sacrificing the precision of its bounds during normal steady-state operation.

To further evaluate the robustness of the soft sensing framework, an additional study was conducted by retraining the MO-LSTM without the inclusion of feedstock characteristics. Detailed outputs of the results are presented in the Supplementary Materials Appendix D, Figures D4-D8. A comprehensive breakdown of the LSTM architectures, optimal, comparative hyperparameter configurations, and performance metrics across all evaluated frameworks is summarised in **Table 3**.

**Table 3:** Summary of Model Performance with Architecture and Training Data Source

| Training Data Source | MO-LSTM Architecture & Hyperparameters * | Target | Validation (mean ± SD) | | | Testing (mean ± SD) | | |
|---|---|---|---|---|---|---|---|---|
| | | | RMSE | $R^2$ | MAPE (%) | RMSE | $R^2$ | MAPE (%) |
| Real Only | 2 LSTM Layers with 128 and 64 neurons respectively. Dropout Rate: 0.49 Learning Rate: 0.0002 Batch Size: 32 | Biogas | 2995 ± 295.4 | 0.713 ± 0.054 | 14.39 ± 1.725 | 2487 ± 152.5 | 0.373 ± 0.074 | 7.286 ± 0.532 |
| | | pH | 0.134 ± 0.002 | 0.123 ± 0.030 | 1.22 ± 0.023 | 0.127 ± 0.009 | -0.639 ± 0.241 | 1.306 ± 0.087 |
| | | FOS/TAC | 0.077 ± 0.005 | -0.097 ± 0.138 | 22.03 ± 1.292 | 0.11 ± 0.007 | -0.514 ± 0.196 | 32.14 ± 1.981 |
| Real Only (Optuna based Model Hyperparameters) | 1 LSTM Layer with 128 neurons. Dropout Rate: 0.397 Learning Rate: 0.007 Batch Size: 32 | Biogas | 1902 ± 102.2 | 0.7369 ± 0.0282 | 6.057 ± 0.3401 | 2265 ± 99.75 | 0.4808 ± 0.0456 | 6.334 ± 0.2911 |
| | | pH | 0.1438 ± 0.003 | 0.094 ± 0.034 | 1.2734 ± 0.074 | 0.096 ± 0.002 | 0.077 ± 0.032 | 0.8995 ± 0.0278 |
| | | FOS/TAC | 0.076 ± 0.002 | -0.008 ± 0.06 | 19.04 ± 0.6373 | 0.096 ± 0.002 | -0.1284 ± 0.038 | 28.75 ± 0.811 |
| LSTM-β-VAE Augmented (Optuna based Model Hyperparameters from Real Data Configuration) | 1 LSTM Layer with 128 neurons. Dropout Rate: 0.397 Learning Rate: 0.007 Batch Size: 32 | Biogas | 1113 ± 150 | 0.9088 ± 0.023 | 3.373 ± 0.5261 | 1327 ± 58.21 | 0.838 ± 0.014 | 3.687 ± 0.2737 |
| | | pH | 0.067 ± 0.003 | 0.805 ± 0.02 | 0.5917 ± 0.026 | 0.05 ± 0.003 | 0.7069 ± 0.041 | 0.493 ± 0.021 |
| | | FOS/TAC | 0.04 ± 0.002 | 0.7237 ± 0.025 | 10.42 ± 0.8203 | 0.0493± 0.004 | 0.738 ± 0.049 | 13.78 ± 1.153 |
| LSTM-β-VAE Augmented (Optuna based Model Hyperparameters) | 3 LSTM Layers with 128,128 and 64 neurons respectively. Dropout Rate: 0.325 Learning Rate: 0.001 Batch Size: 16 | Biogas | 1257± 156.1 | 0.8838 ± 0.03 | 3.679 ± 0.2881 | 1180 ± 79.98 | 0.8715 ± 0.017 | 3.302 ± 0.1961 |
| | | pH | 0.058 ± 0.004 | 0.8544 ± 0.018 | 0.5072 ± 0.022 | 0.0343 ± 0.003 | 0.8591 ± 0.027 | 0.3048 ± 0.03 |
| | | FOS/TAC | 0.031 ± 0.0002 | 0.8365 ± 0.003 | 7.775 ± 0.2601 | 0.0327 ± 0.003 | 0.8851 ± 0.017 | 9.219 ± 0.5039 |
| LSTM-β-VAE Augmented (without Feedstock Characteristics) | 3 LSTM Layers with 256, 128 and 16 neurons respectively. Dropout Rate: 0.346 Learning Rate: 0.003 Batch Size: 64 | Biogas | 927.5 ± 43.9 | 0.9375 ± 0.006 | 2.803 ± 0.096 | 1008 ± 55.67 | 0.9063 ± 0.0100 | 2.842 ± 0.1718 |
| | | pH | 0.067 ± 0.003 | 0.8013 ± 0.017 | 0.4974 ± 0.043 | 0.036 ± 0.004 | 0.8437 ± 0.033 | 0.3091 ± 0.021 |
| | | FOS/TAC | 0.034 ± 0.003 | 0.7921 ± 0.036 | 8.741 ± 0.6585 | 0.038 ± 0.003 | 0.8486 ± 0.022 | 10.9 ± 0.6974 |
| LSTM-β-VAE | 1 LSTM Layer with 128 | Biogas | 923.8 ± 69.71 | 0.9378 ± 0.01 | 2.663 ± 0.2441 | 1028± 25.31 | 0.9029 ± 0.005 | 2.917 ± 0.088 |
| Augmented (without Feedstock Characteristics but (Optuna based Model Hyperparameters from Real Data Configuration) | neurons. Dropout Rate: 0.397 Learning Rate: 0.007 Batch Size: 32 | pH | 0.075 ± 0.003 | 0.7554 ± 0.017 | 0.5731 ± 0.048 | 0.044 ± 0.006 | 0.7616 ± 0.06 | 0.4192 ± 0.042 |
| | | FOS/TAC | 0.037 ± 0.002 | 0.7576 ± 0.028 | 9.157 ± 0.7721 | 0.04 ± 0.002 | 0.8327 ± 0.016 | 11.3 ± 0.5748 |

* All models were trained using the Adam optimiser for a maximum of 200 epochs, with an early-stopping patience of 30 epochs.

The exclusion of these variables did not degrade predictive accuracy but rather resulted in marginally improved overall performance. The newly optimised multi-output LSTM (consisting of three layers with 256, 128, and 16 units) achieved test $R^2$ values of 0.91, 0.84, and 0.85 for Total Biogas, pH, and FOS/TAC, respectively. While biogas predictive capabilities slightly increased, pH and FOS/TAC estimation experienced only negligible declines which may attribute to high correlation of the feedstock characteristic information on stability indicators. This can be attributed to the specific operational profile of the studied site, which operates on a relatively homogenous and stable agricultural diet (primarily maize and rye silage with occasional grass silage). Consequently, forcing the network to map these noisy, low-variance inputs increases the dimensionality of the feature space without enhancing their predictive capability unless they introduce new temporal patterns [34].

While this finding demonstrates that predictive monitoring can successfully operate at this specific site without the financial and logistical burden of feedstock characterisation for new for every delivery, the result must be carefully contextualised. For sites processing highly variable and complex co-digested feeds dynamic feedstock characterisation may remain important. Therefore, further investigation across diverse operational regimes is required before generalising this finding.

### 3.3 Effect of Synthetic Training-Set Size

Using synthetic data generated from the LSTM-β-VAE has shown to improve the generalisability for the downstream M0-LSTM. The size of the synthetic training set was initially set a at 5,000 sequences. This is roughly an order of magnitude larger than the 366 daily windows available from one year of real training data and was chosen to give dense coverage of the VAE's learned distribution. Predictive accuracy improved steadily with the number of synthetic training sequences as presented in **Table 4**. With 366 and 500 sequences, the MO-LSTM performed no better than the real-data baselines in **Table 2**. The benefit of augmentation therefore comes from the larger number and greater diversity of trajectories the LSTM-β-VAE can produce, not from the synthetic samples being cleaner than real data.

**Table 4:** Effect of synthetic training-set size on MO-LSTM test performance (fixed hyperparameters, mean ± SD over five seeds).

| Training sequences | Biogas R² | Biogas MAPE (%) | pH R² | pH RMSE | FOS/TAC R² | FOS/TAC MAPE (%) | Mean R² |
|---|---|---|---|---|---|---|---|
| 366 | 0.15 ± 0.06 | 8.89 ± 0.54 | 0.05 ± 0.01 | 0.089 ± 0.000 | 0.11 ± 0.03 | 25.5 ± 0.2 | 0.10 ± 0.02 |
| 500 | 0.27 ± 0.03 | 8.10 ± 0.23 | 0.08 ± 0.01 | 0.088 ± 0.001 | 0.13 ± 0.04 | 25.3 ± 1.8 | 0.16 ± 0.02 |
| 1,000 | 0.76 ± 0.04 | 4.67 ± 0.51 | 0.57 ± 0.12 | 0.060 ± 0.008 | 0.75 ± 0.02 | 13.6 ± 0.9 | 0.69 ± 0.05 |
| 2,500 | 0.86 ± 0.02 | 3.63 ± 0.24 | 0.77 ± 0.04 | 0.043 ± 0.004 | 0.88 ± 0.01 | 9.5 ± 0.4 | 0.84 ± 0.01 |
| 5,000 | 0.87 ± 0.02 | 3.30 ± 0.19 | 0.85 ± 0.02 | 0.035 ± 0.002 | 0.88 ± 0.02 | 9.3 ± 0.6 | 0.87 ± 0.02 |

The largest change occurred between 500 and 1,000 sequences, where mean test $R^2$ rose from 0.16 to 0.69. The stability indicators gained the most: FOS/TAC $R^2$ rose from 0.13 to 0.75 and pH from 0.08 to 0.57. Beyond 1,000 sequences, returns decreased. Going from 2,500 to 5,000 sequences raised mean test $R^2$ by only 0.03, from 0.84 to 0.87. Most of that remaining gain came from pH (0.77 to 0.85), which was the slowest output to converge and the most variable across seeds at intermediate sizes (± 0.12 at N1000). Seed-to-seed variability also fell as data increased. Together with the falling validation loss (0.44 to 0.24), this indicates that the larger synthetic sets regularised the network rather than simply extending training.

Re-tuning hyperparameters at each size presented in the Supplementary Materials Appendix D, Figure D9-D10 and **Table D1** gave the same overall pattern (mean test $R^2$ from 0.12 to 0.85), which confirms that the trend does not depend on the fixed architecture. At intermediate sizes, per-size tuning was generally less stable than the fixed configuration. For example, Total Biogas $R^2$ at N2500 was 0.74 ± 0.07 with tuning, compared with 0.86 ± 0.02 without it. This is consistent with the hyperparameter search itself overfitting the validation set when only 30 trials are used on smaller datasets.

However, computational cost grew roughly linearly with data size. Total Optuna time rose from 0.09 h (N366) to 1.16 h (N5000), and final training time from about 10 s to about 350 s. Inference time stayed below 1 ms per sequence at every size, where it depended on the selected architecture rather than on the volume of training data. For example, the N1000 model had about 328,000 parameters compared with 40,000 at N2500. So larger synthetic sets add a one-off offline training cost but no deployment cost. As a trade-off, 2,500-5,000 sequences were the most practical operating range for this site. Further gains beyond 5,000 are most likely to be small, and the limit will probably be set by the diversity of the VAE's learned distribution rather than by sample count.

The quality and physical reliability of the synthetic data inherently determine the subsequent prediction performance, particularly because industrial AD datasets consist primarily of stable, normal operating conditions. When the original dataset contains very limited examples of abnormal operating conditions, the predictive model is entirely reliant on the generative model to define the boundaries of process failure. If the synthetic sequences were of poor quality, such as failing to capture the full operational distribution or producing chemically invalid transitions, the downstream predictive model would learn inaccurate process dynamics. In a practical setting, this would cause the predictive tool to generate false instability warnings or entirely miss critical failure events. Since the proposed generative framework successfully reproduces the dynamic volatility and limits of the real system, it enables the predictive model to learn accurate transition boundaries rather than overfitting to historical steady-state means. For practical applications, this demonstrates that reliable early-warning models can be deployed at AD sites constrained by sparse failure data, provided the augmentation process strictly preserves temporal and multivariate fidelity. To ensure the model remains accurate over the long term, operators would need to periodically retrain the generative model as new extreme operational deviations occur, ensuring the synthetic distributions do not drift from the evolving physical reality of the digester.

## 4.0 Conclusions

This study developed a multi-output LSTM framework for full-scale AD, addressing industrial data sparsity and the joint estimation of production and biological stability indicators. Baseline evaluations demonstrated that sparse historical transition states caused standard tabular algorithms and unaugmented deep learning architectures failed to reproduce transient deviations. Introducing an LSTM-β-VAE augmentation pipeline populated sparse regions of the training space with synthetic sequences that retained the correlation structure of the measured data. Trained on this augmented dataset, the MO-LSTM achieved robust simultaneous estimation (test $R^2$ 0.87-0.89) across all key outputs. A training-size study further showed that accuracy depended on the number of synthetic sequences, rising sharply between 500 and 1,000 sequences and plateauing between 2,500 and 5,000. Furthermore, integrating locally adaptive conformal prediction provided rigorous uncertainty quantification, dynamically signalling increased uncertainty during abnormal biochemical shifts within well-calibrated 90% coverage intervals. A streamlined model relying exclusively on standard online SCADA parameters removing the need for offline feedstock characterisation achieved slightly improved predictive capabilities ($R^2$ 0.85-0.91).

Despite these promising outcomes, several limitations must be acknowledged. First, the dataset was restricted to a single site with a relatively small sample size of unique operational events. Second, due to the infrequent nature of offline laboratory measurements, the framework relied on interpolated data for the critical stability indicators, which may mask underlying high-frequency biological transients. Finally, the framework currently lacks external validation. Its ability to generalise to other digesters with different physical configurations remains untested.

While the framework demonstrates strong adaptability, several areas require further investigation. Future work must address the fixed 30 day step look-back window by developing dynamic optimisations to better capture the delayed degradation kinetics of variable seasonal feedstocks. To validate the generalisability of this framework across diverse operational environments, future studies should investigate transfer learning approaches. This would allow the pre-trained model to be rapidly adapted to other full-scale AD facilities requiring only minimal site-specific data for fine-tuning.

Additionally, to build upon the generative pipeline, future studies should conduct comprehensive comparative benchmarking of various time-series augmentation techniques. While the LSTM-β-VAE preserved statistical manifolds, purely data-driven generative models do not inherently enforce physical constraints. Future research must explicitly validate synthetic transition states against deterministic biochemical bounds and verifying the mass balances, ensuring specific biogas yields remain within theoretically achievable limits.

Finally, to overcome the extrapolation limits and black box nature of purely data-driven LSTMs, future work should explore hybrid modelling. By combining mechanistic knowledge of the AD process with AI, the framework will be constrained by fundamental physics and biology. Such an approach would be a prerequisite for safe real-time integration into plant SCADA systems as an automated early-warning tool.

**CRediT authorship contribution statement**

Rohit Murali: Writing – original draft, Investigation, Methodology, Software, Visualisation, Conceptualisation. Benaissa Dekhici: Writing – review & editing, Conceptualisation. Tao

Chen: Writing – review & editing, Supervision, Project administration, Funding acquisition. Dongda Zhang: Writing – review & editing, Validation, Supervision, Project administration, Funding acquisition. Michael Short: Writing – review & editing, Supervision, Resources, Project administration, Investigation, Funding acquisition, Conceptualisation.

## Declaration of competing interest

The authors declare that they have no known competing financial interests or personal relationships that could have appeared to influence the work reported in this paper.

## Declaration of generative AI and AI-assisted technologies in the manuscript preparation process

During the preparation of this work, the authors used Google Gemini to improve readability and language. After using this tool/service, the authors reviewed and edited the content as needed and takes full responsibility for the content of the published article.

## Acknowledgements

This work was kindly supported and funded by the Surrey Institute for People-Centred AI (PAI) and by the Engineering and Physical Sciences Research Council EP/Y005600/1 (AI for Net Zero programme). The authors also thank the Supergen Bioenergy Impact Hub, EP/Y016300/1 for support and acknowledge Future Biogas Ltd. For providing support for the research in this paper.

## Data Availability

The authors do not have permission to share commercial confidential data.